\documentclass[letterpaper]{article} 
\usepackage{aaai25}  
\usepackage{times}  
\usepackage{helvet}  
\usepackage{courier}  
\usepackage[hyphens]{url}  
\usepackage{graphicx} 
\usepackage{natbib}  
\usepackage{caption} 
\usepackage{algorithm}
\usepackage{algorithmic}
\usepackage{booktabs}       
\usepackage{amsfonts}       
\usepackage{amsmath}
\usepackage{amssymb}
\usepackage{epsfig}
\usepackage{graphicx}
\usepackage{booktabs}
\usepackage{bbding}
\usepackage{pifont}
\usepackage{booktabs}
\usepackage{graphicx}
\usepackage{multirow}
\usepackage{capt-of}
\usepackage{graphicx, amsmath, amssymb, caption, subcaption, multirow, overpic, textpos}
\usepackage[british, english, american]{babel}
\usepackage{booktabs}       
\usepackage{amsfonts}       
\usepackage{multirow}
\usepackage{amsmath}
\usepackage{newfloat}
\usepackage{listings}
\DeclareCaptionStyle{ruled}{labelfont=normalfont,labelsep=colon,strut=off} 
\floatstyle{ruled}
\newfloat{listing}{tb}{lst}{}
\floatname{listing}{Listing}
\usepackage{algorithm}
\usepackage{algorithmic}
\usepackage{booktabs}       
\usepackage{amsfonts}       
\usepackage{amsmath}
\usepackage{amssymb}
\usepackage{epsfig}
\usepackage{graphicx}
\usepackage{booktabs}
\usepackage{bbding}
\usepackage{pifont}
\usepackage{booktabs}
\usepackage{graphicx}
\usepackage{multirow}
\usepackage{capt-of}
\usepackage{graphicx, amsmath, amssymb, caption, subcaption, multirow, overpic, textpos}
\usepackage[british, english, american]{babel}
\usepackage{booktabs}       
\usepackage{amsfonts}       
\usepackage{multirow}
\usepackage{amsmath}
\usepackage{newfloat}
\usepackage{listings}
\usepackage{romannum}
\DeclareCaptionStyle{ruled}{labelfont=normalfont,labelsep=colon,strut=off} 
\floatstyle{ruled}
\newfloat{listing}{tb}{lst}{}
\floatname{listing}{Listing}
\usepackage{xcolor}
\newlength\savewidth\newcommand\shline{\noalign{\global\savewidth\arrayrulewidth
  \global\arrayrulewidth 1pt}\hline\noalign{\global\arrayrulewidth\savewidth}}
\newcommand{\tablestyle}[2]{\setlength{\tabcolsep}{#1}\renewcommand{\arraystretch}{#2}\centering\footnotesize}
\renewcommand{\paragraph}[1]{\vspace{1.25mm}\noindent\textbf{#1}}

\title{Review Learning: Advancing All-in-One Ultra-High-Definition Image Restoration Training Method}
\author{
    Xin Su$^{1}$,~~ 
    Zhuoran Zheng$^{2}$\thanks{$*$ indicates corresponding author},~~
    Chen Wu$^{3}$
}
\affiliations{
    \textsuperscript{\rm 1}Fuzhou University
    \\
    \textsuperscript{\rm 2*}Sun Yat-sen University
    \\
    \textsuperscript{\rm 3}University of Science and Technology of China
    \\
    suxin4726@gmail.com, zhengzr@njust.edu.cn, wuchen5X@mail.ustc.edu.cn

}

\begin{document}
\twocolumn[{%
\renewcommand\twocolumn[1][]{#1}%
\maketitle
\begin{center}\footnotesize
    \centering
    \captionsetup{type=figure}
    \includegraphics[width=18cm]{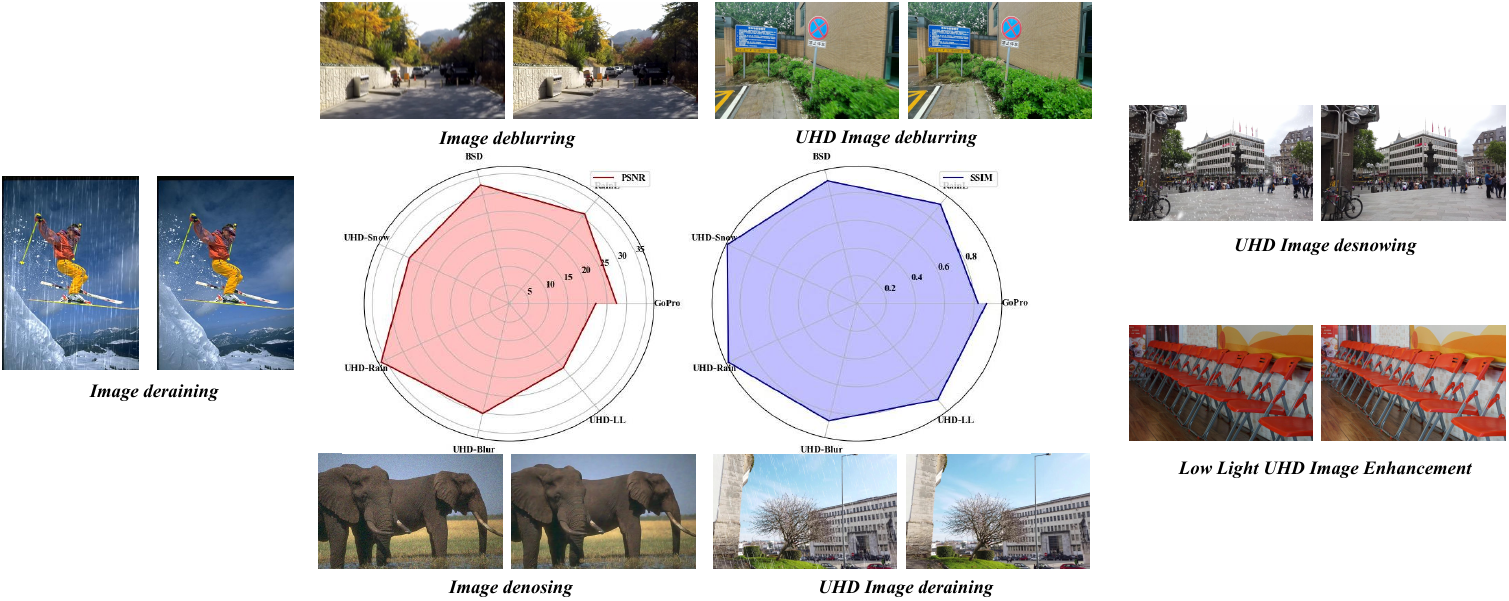}
    \captionof{figure}{This paper leverages the \textbf{Review Learning} training method for an all-in-one image restoration task. 
    In contrast to existing all-in-one modeling methods for image restoration, we only leverage customized training schemes to optimize the image restoration network without a priori and prompting information.
    Our method achieves insightful performance on seven benchmarks (including degraded UHD images and low-resolution images) compared to other algorithms.}
    \label{fig:com1}
\end{center}%
}]
\begin{abstract}
All-in-one image restoration tasks are becoming increasingly important, especially for ultra-high-definition (UHD) images. 
Existing all-in-one UHD image restoration methods usually boost the model's performance by introducing prompt or customized dynamized networks for different degradation types.
For the inference stage, it might be friendly, but in the training stage, since the model encounters multiple degraded images of different quality in an epoch, these cluttered learning objectives might be information pollution for the model.
To address this problem, we propose a new training paradigm for general image restoration models, which we name \textbf{Review Learning}, which enables image restoration models to be capable enough to handle multiple types of degradation without prior knowledge and prompts.
This approach begins with sequential training of an image restoration model on several degraded datasets, combined with a review mechanism that enhances the image restoration model's memory for several previous classes of degraded datasets.
In addition, we design a lightweight all-purpose image restoration network that can efficiently reason about degraded images with 4K ($3840 \times 2160$) resolution on a single consumer-grade GPU. 

\end{abstract}
\section{Introduction}
In recent years, the emergence of sophisticated imaging sensors and displays has greatly contributed to the progress of Ultra-High-Definition (UHD) imaging.
However, the increased number of pixels in UHD images undoubtedly makes them more susceptible to multiple unknown degradations during the imaging process.
Previous UHD image restoration methods often address these basic image degradations separately, including low-light image enhancement~\cite{Li2023ICLR}, deraining~\cite{7893758, 8099669, Fu_Qi_Zha_Zhu_Ding_2022,8953954}, debluring, and dehazing~\cite{Cai_Xu_Jia_Qing_Tao_2016,5567108}, by using specific single-task models.
Indeed, there is a greater need for an all-in-one model to restore different types of UHD degraded images.
\par A recent line of work~\cite{li2022all,potlapalli2023promptir} has focused on unified image restoration, where models are trained on a mixed degradation dataset and implicitly classify the type of degradation in the restoration process.
While the results are impressive, the reduction of batch size to fit consumer-grade GPUs' memory constraints leads to prolonged training times.
In addition, this training method relies on prompt signals with strong perceptual capabilities, unfortunately, these prompts struggle to capture the degraded type of image.
In light of this, we ask two key questions for the all-in-one image restoration task:

\noindent 
\textcolor{red}{i)} \textit{Can we optimize the model by mixed degradation images without prompts?} 

\noindent \textcolor{red}{ii)} \textit{How to avoid information pollution (interference of multiple degraded types of images on model memory) during the training model phase?}

\par To address these issues, and inspired by the principles of Continual Pre-Training (CPT) as demonstrated in recent studies~\cite{law_llm,med_llm,cpt_llm_survey,empirical_cv_catastrophic_forgetting}, we propose an approach named Review Learning that incrementally enhances arbitrary model's capabilities in unified image restoration areas without incurring a significant general performance penalty.
Specifically, the model learns a specific task (such as low light image enhancement, or image deblurring) and then builds upon this knowledge to learn new tasks, with regular reviews of challenging samples from previous tasks to avoid knowledge loss.
We perform comprehensive analyses and statistics to determine the frequency and intervals for reviewing difficult samples.
This paradigm for training models exploits the maximum potential of neural networks without prompting information.
The key findings of this paper can be summarized as follows:
\begin{itemize}
    \item To the best of our knowledge, we propose the review learning method that is the first all-in-one image restoration algorithm that does not require a priori knowledge and does not require a customized network architecture, which gives the network the ability to solve multiple degradation types through an iterative learning strategy.
\end{itemize}
\begin{itemize}
    \item We propose a lightweight efficient general model for image restoration called SimpleIR. SimpleIR can perform full-resolution inference of UHD images (ie. 4K, 8K) on a single consumer-grade GPU.
\end{itemize}
\begin{itemize}
    \item Extensive comparisons validate the effectiveness of SimpleIR by achieving state-of-the-art performance on various image restoration tasks, including LLIE, image deraining, deblurring, and desnowing.
\end{itemize}
\begin{figure}[!t]
\setlength{\abovecaptionskip}{0.1cm} 
\setlength{\belowcaptionskip}{-0.5cm}
	\centering
	\includegraphics[width=\linewidth]{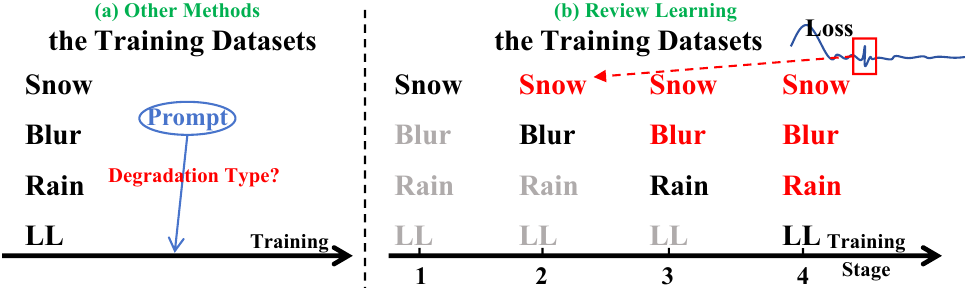}\\
	\caption{We propose a review learning method that trains an end-to-end network sequentially on different types of datasets. It is worth noting that to avoid catastrophic forgetting of the model, our network is trained on a new dataset by reviewing previously trained difficult samples of other degenerate types. These difficult samples are stored by observing fluctuations in the loss during training.}
	\label{fig:arch}
\end{figure}
\section{Related work and Motivation}
\noindent\textbf{Unified All-in-One Image Restoration} 
A body of recent works pays attention to unified image restoration where multiple image restoration tasks are handled with a single model. 
They archive impressive results by using a prompt module~\cite{9157460,potlapalli2023promptir,10378570} or applying pre-trained VLMs~\cite{luo2023controlling} to distinguish between different degradation types to avoid information pollution.
However, the former methods are still limited to the hand-crafted nature of these prompts and may not fully capture the underlying patterns and structures present in the data. On the other hand, while VLMs offer a more flexible and data-driven approach to handling various unknown degradation types, they still face challenges related to the `information pollution' phenomenon, which can impair the model's ability to discern subtle degradation features. This occurs when the model struggles to maintain a clear and distinct representation of each degradation type in the presence of diverse and potentially conflicting information within the training data. \\
\noindent \textbf{UHD Images Restoration} The quest for UHD image restoration has surged in recent years, driven by the demand for enhanced visual fidelity in various applications \cite{zhang2018image, 9156921, 9578037}. Most existing works train multiple models for different tasks separately and focus on addressing one degradation type at a time and do not explore the potential of handling multiple degradation types simultaneously, such as multi-guided bilateral learning for UHD image dehazing  \cite{zheng2021ultra}, Fourier embedding network UHDFour~\cite{Li2023ICLR_uhdfour} for UHD low-light image enhancement, and explore the feature transformation from high-resolution and low-resolution for UHD image dehazing \cite{wang2023ultra}. Predictably, building an end-to-end lightweight model is needed for UHD images.

\noindent \textbf{Motivation}
i) The drawbacks of prompt-based approaches: Traditional methods that rely on specific prompts limit the model's ability to generalize across unknown types of image degradation.  In addition, the generation of prompts usually requires a large language model, which is often costly.
ii) Propomt information generative and dynamically structured networks tend to hallucinate when confronted with images of unknown degradation or compound degradation types.
\begin{algorithm}[!t]
\footnotesize
    \caption{Implementing Review Learning}
    \label{alg:1}
    \begin{algorithmic}
        \REQUIRE the N types of degradation UHD datasets $D_{i} (I = 1, 2, 3,..., n)$ and the ranking of the entropy difference of datasets $R_{j} (j = 1, 2, 3,..., n)$. 
        \ENSURE the unified set of trained weights. 
        \vspace{0.5mm}
        \STATE\hspace*{-2mm}\textbf{Procedure:} 
        \vspace{1.5mm}
        \STATE Initialize the model parameters $\theta$.
        \vspace{1mm}
    \FOR{each dataset $D_{i}$ in the order of $R_{j}$}
        \STATE Train the model on the current dataset $D_{i}$ to convergence.
        \IF{dataset $D_{i}$ is the second dataset in the order}
            \STATE Calculate the average loss $\mu$ and standard deviation $\sigma$ of the current training set.
            \STATE $S_{loss} \gets$ samples with loss greater than $\mu + \kappa\sigma$ where $\kappa$ is a constant.
            \STATE Mix samples from $S_{loss}$ into the training set $D_{i}$.
        \ENDIF
        \STATE Collect challenging samples based on high entropy differences.
        \STATE Update the model parameters $\theta$ with the learned weights.
    \ENDFOR
    \vspace{1mm}
    \FOR{each subsequent dataset $D_{i+1}$}
        \STATE Fine-tune the model on the new dataset $D_{i+1}$ with parameters $\theta$.
        \STATE Mix a portion of previously collected challenging samples into the training set.
        \STATE Update the model parameters $\theta$.
    \ENDFOR  
    \vspace{1mm}
    \RETURN the final set of pre-trained weights $\theta$.
    \end{algorithmic}
\end{algorithm}
\section{Methodology}
We propose `Review Learning' which aims at finding a unified set of pre-trained weights to archive universal image restoration.
In our training paradigm, the challenging samples from the previous training stage will be reviewed in the subsequent stage. 
Moreover, the number of challenging samples is significantly smaller than the main training dataset. This ensures that, while avoiding information pollution, we can also effectively prevent catastrophic forgetting.
In this section, we will elaborate on `Review Learning' and the lightweight general UHD image restoration model.
\subsection{Review Learning}
\noindent \textbf{Theoretical Foundation}
At the core of our `Review Learning' methodology lies the concept of continual learning. This theory suggests that an effective learning system should incrementally accumulate knowledge, akin to human learning, where new information is absorbed without eroding the recollection of prior experiences~\cite{parisi2019continual}. 
In neural network paradigms, this is often complicated by the phenomenon of catastrophic forgetting~\cite{french1999catastrophic}, where the network's ability to accommodate new data can come at the expense of its stored memories.
So far, we have designed a review learning paradigm by combining the idea of step-by-step learning and strategies to avoid catastrophic forgetting.
\noindent \textbf{Design of Review Learning}
As shown in Algorithm~\ref{alg:1}, the whole implementing process of Review Learning includes three
steps, which are introduced as follows:
\begin{enumerate}
    \item \textbf{Training Pipeline Based on Degradation Complexity:} 
    Ranking the training order of degradation datasets based on their intrinsic learning difficulty, which is determined by the entropy difference between original and degraded images. This ranking facilitates a structured learning approach, starting from the simplest to the most complex tasks.
    \item \textbf{Acquisition of Challenging Samples:}   
    In the initial phase, challenging samples are identified by their loss values that exceed the local mean, indicating the difficulty of the model. From the second phase onwards, the focus shifts to samples with the highest entropy differences, avoiding the mixed dataset distribution, which could misrepresent the true challenging samples.
    \item \textbf{Sequential Training and Review Integration:} The model undergoes sequential training on datasets, starting afresh with the first. After each phase, challenging samples are identified and archived. As training advances to subsequent datasets, a selection of these challenging samples is reintegrated, ensuring continuous reinforcement of learned knowledge and mitigation of catastrophic forgetting.
\end{enumerate}
%
\par
\begin{figure}[htbp]\footnotesize
    \centering
    \includegraphics[width=1\linewidth]{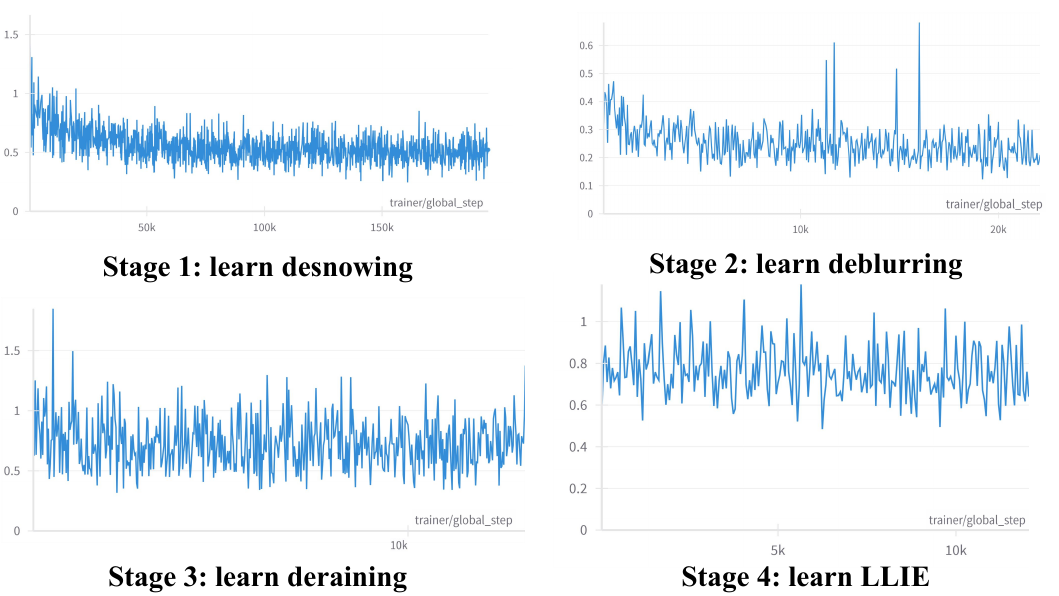}
    \caption{The training loss curve illustrates the model's performance across various stages of multi-degradation task learning.}
    \label{fig:loss}
    \vspace{-1mm}
\end{figure}

 In this section, we elaborate on the restoration process for four types of degraded UHD images to help explain the effectiveness of `Review Learning'. 
\begin{figure}[!t]\footnotesize
    \centering
    \includegraphics[width=1\linewidth]{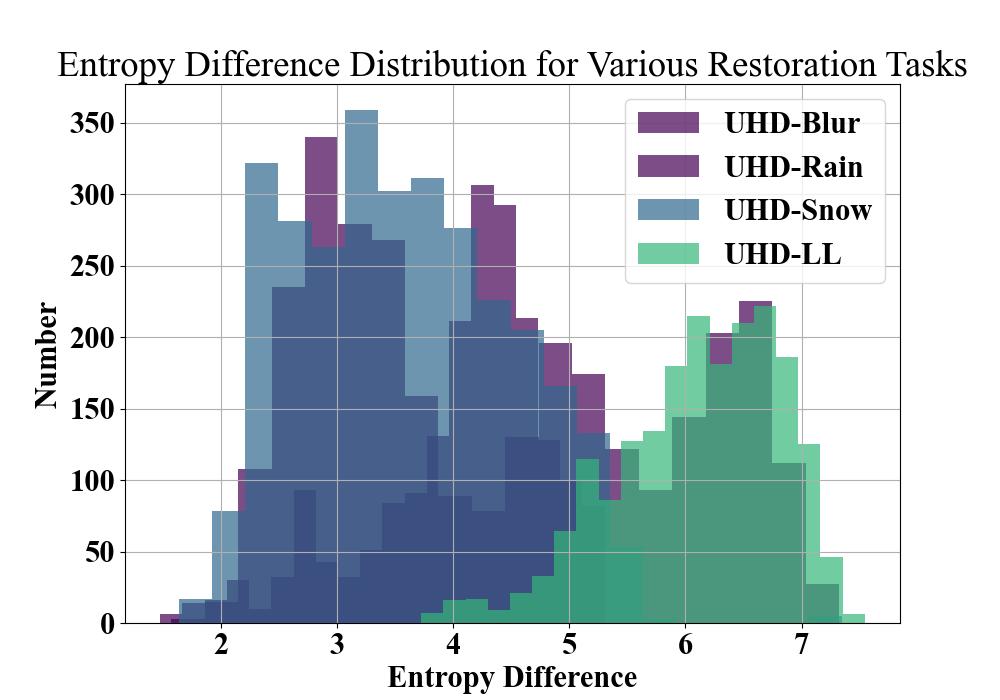}
    \caption{Histograms of entropy differences for 4 image degradation datasets.}
    \label{fig:entropy}
    \vspace{-5mm}
\end{figure}
As depicted in Figure~\ref{fig:entropy}, we show a histogram of entropy differences across four types of UHD-degraded datasets. We strategically plan the learning pathway for the model based on the difficulty level, which progresses from desnowing to deblurring, deraining, and finally to low-light enhancement.
Initially, we train the model on the UHD-Snow dataset and collect challenging samples based on abnormally high loss values that exceed the average loss during the surrounding epochs.
After the pre-training (Stage 1), we select all challenging samples, accounting for 10\% of the UHD-Snow dataset.
In Figure~\ref{fig:loss}, it is observed that the training loss fluctuates violently after Stage 1. So we select the challenging samples with the top 20\% entropy differences, irrespective of their loss function performance.
In addition, we reduce the number of the previous-phase difficult samples reviewed by 50\% in each subsequent phase to avoid `information pollution`.
%
\subsection{SimpleIR Architecture}
For a given degraded image \(\textbf{I}\in \mathbb{R}^{H\times W\times3}\), we first apply a $3$$\times$$3$ convolution and a sub-pixel convolutional layer to obtain the shallow feature $\mathbf{X}_{\text{0}}$~$\in$~$\mathbb{R}^{\frac{H}{4} \times \frac{W}{4} \times C}$; where $H\times W$ denotes the spatial dimension and $C$ is the number of channels. 
Then, the multiple stacked Feature Iteration Blocks (FIBs) are used to generate finer
deep features $\mathbf{X}_{\text{i}}$ from $\mathbf{X}_{\text{0}}$ for degraded image reconstruction, where a FIB has a Hybrid Attention Block (HAB) and a
Feed-forward Network (FFN). Lastly, the sum of the final features $\mathbf{X}_{\text{f}}$
and $\mathbf{X}_{\text{0}}$ is fed to a $3$$\times$$3$ convolution and a sub-pixel convolutional layer obtains the restored image.
See the Appendix for the overall framework of our SimpleIR.
\subsubsection{Hybrid Attention Blocks}
To capture low-resolution features of images, we propose the Hybrid Attention Block (HAB) to endow the low-resolution space with more representative features. 
Each HAB contains a DualStream Attention (DSA) and a Local Detail Attention Module (LDAM). Here, we describe our HAB pipeline in general, which can be
written as:

\begin{equation}
\mathbf{\hat{X}_{\text{in}}}=\mathrm{Conv}[\mathrm{DSA}(\mathrm{LN}(\mathbf{{X}_{\text{in}}}));\mathrm{LDAM}(\mathrm{LN}(\mathbf{{X}_{\text{in}}}))],
\end{equation}

\noindent where \(X_{in}\) stand for the input features, \(\hat{X}_{in}\) are the intermediate features. [;] is a concatenation operation, and Conv represents the 1 × 1 convolution.

\noindent\textbf{DualStream Attention}
The major computational overhead in the model comes from the self-attention layer. Numerous approaches have recently been proposed to reduce the quadratic complexity of self-attention, the computational bottleneck in transformer-based architectures.
In the quest for efficient and scalable attention mechanisms suitable for real-time mobile vision applications, we introduce a novel approach by opting instead for a more computationally efficient strategy that retains the essence of capturing global dependencies within the input data.
At the core of our DualStream Attention lies a dual pathway for the generation of query and key vectors, designed to encapsulate the cross-channel and spatial context of the input feature maps. 
We employ a series of convolutional operations, including 1$\times$1 and 3$\times$3 convolutions, to aggregate pixel-wise and channel-wise information, respectively.
This results in the formulation of query $\mathbf{Q}$ and key $\mathbf{K}$ matrices, which are then normalized to ensure consistency in the attention weighting process. This procession can be described as:
\begin{equation}\begin{array}{l}\mathbf{Q} = \mathbf{W_d^Q}\cdot \mathbf{W_p^Q}\cdot \mathbf{X} ,\\\mathbf{K} = \mathbf{W_d^K}\cdot \mathbf{W_p^K} \cdot \mathbf{X}, \end{array}\end{equation}
\noindent where $\mathbf{W_d^{(\cdot)}}$ denotes the 1$\times$1 point-wise convolution, $\mathbf{W_p^{(\cdot)}}$ represents the 3$\times$3 depth-wise convolution, and $X$ is the shallow feature.
%
\par To derive a global context descriptor \(G\in\mathbb{R}^{B\times C}\), we apply the adaptive average pooling to $Q$, which collapses the spatial dimensions while retaining the channel information.
$q$ is then a series of operation layers to produce a channel-wise attention weight $W$:
\begin{equation}
\mathbf{W} = \mathrm{\sigma(FC_2(ReLU(FC_1(GAP(q)))))},
\end{equation}
where $\mathrm{GAP}(\cdot)$ denotes global average pooling, and $FC_{i}$ denotes the full connectivity layer.
The attention weights $W$ are reshaped to match the dimensions of $Q$ and are used to modulate it, emphasizing the most relevant features across the channel dimension:
\begin{equation}
    \mathbf{G} = \mathbf{W} \odot \mathbf{Q}.
\end{equation}
\par The output of the DualStream Attention \(\mathbf{\tilde{X}}\) can be described as:
\begin{equation}
\mathbf{\tilde{X}}  =\mathrm{Conv_{1\times1}}(\mathbf{G} \odot \mathbf{K}) + \mathbf{Q}.
\end{equation}
\noindent\textbf{Local Detail Attention Module} 
Inspired by \cite{7780677} design that efficiently captures information at various scales, we introduce a lightweight Local Detail Attention Module (LDAModule) to enhance the extraction of fine-grained features in high-resolution imagery.

Given an input feature map X, we first split it into four distinct branches, each focusing on different aspects of the feature spaces along the channel dimension:
\begin{equation}
    \mathbf{X} =[\mathbf{X}_{{\mathrm{hw}}},\mathbf{X}_{{\mathrm{w}}},\mathbf{X}_{{\mathrm{h}}},\mathbf{X}_{{\mathrm{id}}}],
\end{equation}
\noindent where $\mathbf{X}_{{\mathrm{hw}}}$ undergoes a depth-wise convolution with a small square kernel to capture local horizontal and vertical details, expressed as \(\mathbf{X}_{\mathrm{hw}}^{\prime}=\mathrm{DWConv}_{\mathbf{k_s\times k_s}}(\mathbf{X}_{\mathrm{hw}})\), $\mathbf{X}_{\mathrm{w}}$ is processed by a depth-wise convolution with a vertical band kernel to detect horizontal edges and patterns, given by \(\mathbf{X}_\mathrm{w}^{\prime}=\mathrm{DWConv}_{\mathrm{1}\times \mathrm{k_b}}(\mathbf{X}_\mathrm{w})\),  $\mathbf{X}_{\mathrm{h}}$ is filtered through a depth-wise convolution with a horizontal band kernel to extract vertical features, denotes as \(\mathbf{X}_\mathrm{h}^{\prime}=\mathrm{DWConv}_{k_b\times1}(\mathbf{X}_\mathrm{h})\), and $\mathbf{X}_{\mathrm{id}}$ is passed through an identity operation to retain the original high-resolution information without alteration, resulting in $\mathbf{X}_{{\mathrm{id}}}^{\prime} = \mathbf{X}_{\mathrm{id}}$.
The outputs of these branches are then intelligently concatenated to enrich the feature representation, ensuring that both local details and global context are preserved:
\begin{equation}
    \mathbf{X}_{{\mathrm{output}}}=\mathrm{Concat}(\mathbf{X}_{{\mathrm{hw}}}^{\prime},\mathbf{X}_{{\mathrm{w}}}^{\prime},\mathbf{X}_{{\mathrm{h}}}^{\prime},\mathbf{X}_{{\mathrm{id}}}^{\prime})
\end{equation}
\subsubsection{Feed-Forward Network}
To transform features into a compact representation, we introduce the FFN module that consists of a $3\times3$ convolution, a $1\times1$ convolution, and a GELU function. The entire process can be represented as follows:
\begin{equation}
    \mathbf{X}_{out}=\mathrm{Conv}_{1\times1}(\mathrm{GELU}(\mathrm{Conv}_{3\times3}(\mathbf{X_{in}}))),
\end{equation}
\noindent where $\mathbf{X}_{in}$ and $\mathbf{X}_{out}$ are the input features and output features, respectively.

\subsubsection{Feature Iteration Block}
Here, we describe our FIB pipeline in general, which can be written as:
\begin{equation}
    \mathbf{\hat{X}_{in}}=\mathrm{Conv}[\mathrm{DSA}(\mathrm{LN}(\mathbf{X_{in}}));\mathrm{LDAM}(\mathrm{LN}(\mathbf{X_{in}}))]+\mathbf{X_{in}},
\end{equation}
\begin{equation}
    \mathbf{X_{out}}=\mathrm{FFN}(\mathrm{LN}(\mathbf{\hat{X}_{in}}))+\mathbf{\hat{X}_{in}},
\end{equation}
\noindent where $\mathbf{{X}_{in}}$ are the input feature, $\hat{X}_{in}$ are the intermediate features and $\mathbf{{X}_{out}}$ denote the output features. [;] is a concatenation operation, and Conv represents the 1 $\times$ 1 convolution.

\section{Experiments}
We evaluate \textbf{SimpleIR} on benchmarks for four UHD image restoration tasks: \textbf{(a)} low-light image enhancement, \textbf{(b)} image deraining, \textbf{(c)} image desnowing, and \textbf{(d)} image deblurring.

%
\begin{table}[htbp]\footnotesize
\tablestyle{2.5pt}{1}
\begin{tabular}{l|cccccc}
\shline
 Dataset &Training samples & Testing samples & Task
\\\shline
UHD-Snow&3,000 &150 & Desnowing \\
UHD-Blur&1,964 &150 & Deblurring \\
UHD-Rain&3,000 &150 & Deraining \\
UHD-LL  &2,000 &115 & LLIE \\
\shline
\end{tabular}
\caption{Datasets Statistics.}
\vspace{-4mm}
\label{tab: Datasets Statistics.} 
\end{table}
%
%
\noindent \textbf{Datasets} 
We use UHD-LL~\cite{Li2023ICLR_uhdfour} and UHD-Blur~\cite{wang2024uhdformer} to conduct UHD low-light image enhancement and deblurring tasks. 
For image deraining and desnowing, we use the datasets of~\cite{wang2024ultrahighdefinitionrestorationnewbenchmarks}.
The statistics of these $4$ datasets are summarised in Tab.~\ref{tab: Datasets Statistics.}.\\
\noindent \textbf{Implementation Details} 
We implemented all the models with the deep learning framework SimpleIR. 
For preprocessing, we randomly crop the full-resolution 4K image to a resolution of 512 $\times$ 512 as the input and data augmentation by horizontal and vertical random flip. 
We use AdamW optimizer with the initial learning rate $2e^{-4}$. To constrain the training of Simple, we use the same loss function~\cite{Kong_2023_CVPR_fftformer} with default parameters.
In the first step of our method, we train the network on the UHD-Snow dataset for 200,000 iterations and begin collecting the difficult samples from the 100,000 iterations. 
In the second step of our method, we continue to train the network on the UHD-Blur dataset for 10,000 iterations and mix the difficult samples from the first step with the learning rate $1e^{-4}$. The latter step has the same setting.
In addition, we identify and incorporate the top 20\% of samples exhibiting high entropy differences as challenging instances into our dataset from the previous iteration. The rationale behind this selection is to account for variations in data distribution, recognizing that not all instances with a loss value exceeding the local mean are inherently difficult samples. High entropy differences suggest a higher degree of uncertainty in the model's predictions, which may not always correlate with the complexity of the sample.
Specifically, we have reduced the review learning quantity by 50\% in comparison to the previous stage.\\
\noindent \textbf{Evaluation} 
Following~\cite{Li2023ICLR_uhdfour}, we adopt commonly-used IQA PyTorch Toolbox\footnote{https://github.com/chaofengc/IQA-PyTorch} to compute the PSNR~\cite{PSNR_thu} and SSIM~\cite{SSIM_wang} scores of all compared methods and also report the trainable parameters (\textbf{Param}).
Since some methods cannot directly process full-resolution UHD images on one consumer-grade GPU, we have to adopt an additional manner to conduct the experiments.
According to UHDFour~\cite{Li2023ICLR_uhdfour}, resizing (\textbf{RS}) the input to the largest size that the model can handle produces better results than splitting the input into four patches and then stitching the result.
Hence, we adopt the resizing strategy for these methods and report whether models need to resize or not.\\
\noindent \textbf{Comparision approaches.} For all tasks, we compare our SimpleIR (train in a single task) to the prevailing approaches in their respective fields such as UHD~\cite{Zheng_uhd_CVPR21}, Uformer~\cite{9878729}, SFNet~\cite{cui2023selective}, MIMO-Unet~\cite{cho2021rethinking_mimo} and Stripformer~\cite{Tsai2022Stripformer}.
We also compare our method (use the Review Learning) with recent unified approaches: NAFNet,  Restormer, PromptIR, and AirNet. 
\subsection{Main Results\\}
%
\begin{figure*}[!t]
\vspace{-5mm}
\footnotesize
\centering
\begin{center}
\begin{tabular}{ccccccccc}
\hspace{-3mm}\includegraphics[width=1\linewidth]{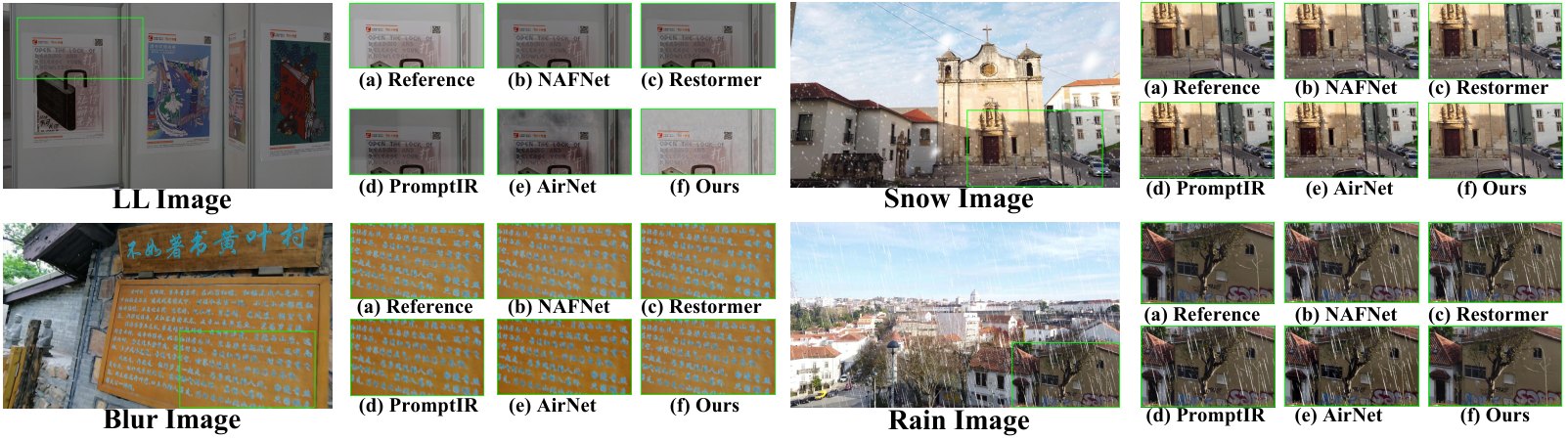}\\ 
\end{tabular}
\vspace{-4mm}
\caption{Qualitative comparisons with NAFNet, Restormer, PromptIR, and AirNet on LLIE, deblurring, deraining, and desnowing tasks.
SimpleIR can generate cleaner results with finer details.
See the Appendix for more results. (Zoom in for a better view)
}
\label{fig:mlti-tasks image enhancement.}
\end{center}
\vspace{-4mm}
\end{figure*}
\begin{figure*}[!t]
\footnotesize
\centering
\begin{center}
\begin{tabular}{ccccccccc}
\hspace{-3mm}\includegraphics[width=1\linewidth]{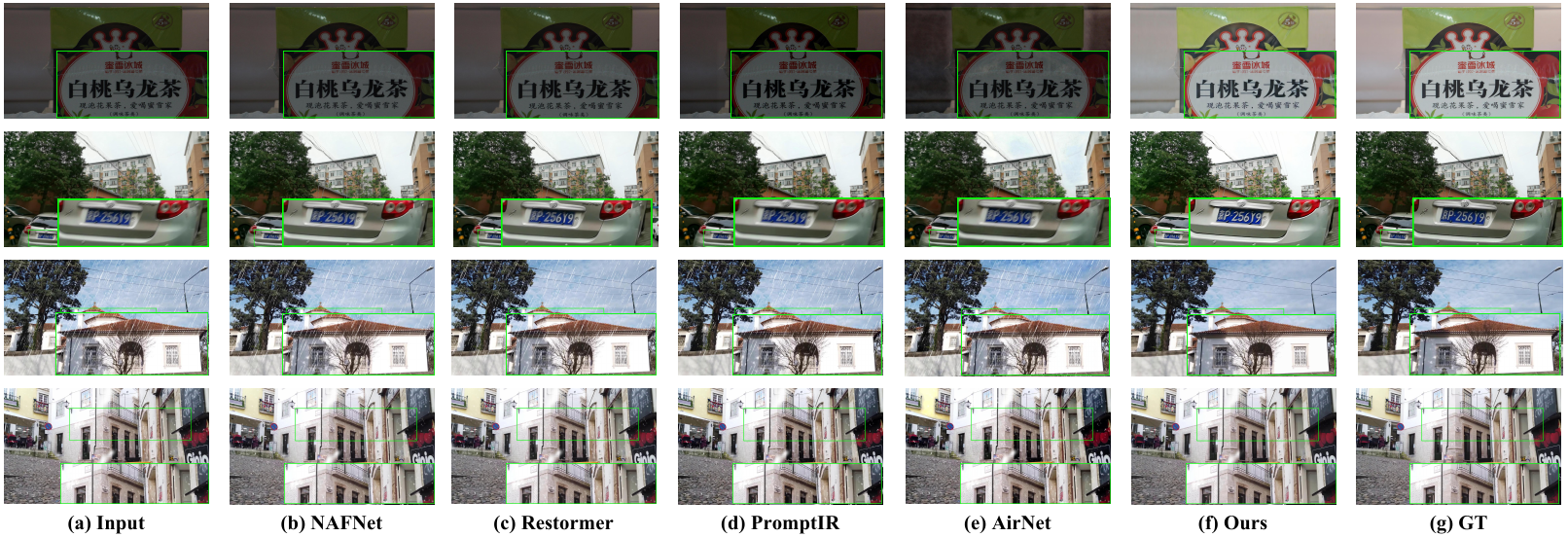}\\ 
\end{tabular}
\vspace{-4mm}
\caption{Visual quality comparisons. The image can be zoomed in for better visualization.}
\label{fig:mlti-tasks image enhancement.}
\end{center}
\vspace{-4mm}
\end{figure*}
\subsubsection{Results on Single-task Image Restoration\\}
\noindent\textbf{Image Desnowing Results} 
We implement a UHD image desnowing experiment on the constructed UHD-Snow dataset.
Tab.~\ref{tab:Image desnowing.} summarises the quantitative results.\\
\noindent\textbf{Image Deblurring Results} 
We evaluate UHD image deblurring with UHD-Blur training sets.
Tab.~\ref{tab:Image deblurring.} summarises the results, where SimpleIR significantly advances current state-of-the-art approaches on different training sets. 
Compared with the recent state-of-the-art deblurring approach FFTformer~\cite{Kong_2023_CVPR_fftformer}, SimpleIR obtains $4.96$dB PSNR gains on UHD-Blur training sets.\\
\noindent\textbf{Image Deraining Results} 
We implement a UHD image deraining experiment on the constructed UHD-Rain dataset.
Tab.~\ref{tab:Image deraining.} summarises the quantitative results. \\
\noindent \textbf{Low-Light Image Enhancement Results} 
We evaluate UHD low-light image enhancement results on UHD-LL training UHD-LL dataset sets.
Tab.~\ref{tab:Low-light image enhancement.} shows that our SimpleIR has impressive results with fewer parameters. 
%
\begin{table}[!t]
\footnotesize
\tablestyle{2pt}{1}
\begin{tabular}{l|c|ccccc}
\shline
 \textbf{Method} &\textbf{Venue} &   \textbf{PSNR}~$\uparrow$ & ~\textbf{SSIM}~$\uparrow$~& ~\textbf{Param}~$\downarrow$~&  \textbf{RS} 
\\
\shline
 \multicolumn{6}{c}{Image desnowing on UHD-Snow}\\
\shline
UHD & ICCV'21&29.29 & 0.950 & 34.5M & \XSolidBrush  \\
Uformer&CVPR'22&23.72 & 0.871 & 20.6M & \CheckmarkBold   \\
SFNet&CVPR'22&23.64 & 0.846&19.7M & \CheckmarkBold  \\
\textbf{SimpleIR (Ours)}&-&\textbf{30.89} & \textbf{0.986} & \textbf{4.6M} &  \XSolidBrush  \\\shline
\end{tabular}
\vspace{-2mm}
\caption{Image desnowing. 
SimpleIR with at least fewer parameters significantly advances state-of-the-art.
}
\label{tab:Image desnowing.} 
\vspace{-2mm}
\end{table}

\begin{table}[!t]
\footnotesize
\tablestyle{2pt}{1}
\begin{tabular}{l|c|ccccc}
\shline
 \textbf{Method} &\textbf{Venue} &   \textbf{PSNR}~$\uparrow$ & ~\textbf{SSIM}~$\uparrow$~& ~\textbf{Param}~$\downarrow$~&  \textbf{RS} 
\\
\shline
 \multicolumn{6}{c}{Image deblurring on UHD-Blur dataset}\\
\shline
MIMO-Unet++&ICCV'21&25.03 &0.752&16.1M & \XSolidBrush   \\
Uformer&CVPR'22&25.26 & 0.752 & 20.6M & 
\CheckmarkBold   \\
Stripformer&ECCV'22&25.05 &0.750&19.7M & \CheckmarkBold  \\
FFTformer&CVPR'23&25.41 & 0.757 & 16.6M & \CheckmarkBold  \\
\textbf{SimpleIR (Ours)}&-&\textbf{31.21} & \textbf{0.871} & \textbf{4.6M} &  \XSolidBrush  \\\shline
\end{tabular}
\caption{Image deblurring. 
SimpleIR with at least fewer parameters significantly advances state-of-the-art.
}
\label{tab:Image deblurring.} 
\vspace{-2mm}
\end{table}

\begin{table}[!t]\footnotesize
\tablestyle{2.2pt}{1}
\begin{tabular}{l|c|ccccc}
\shline
 \textbf{Method} &\textbf{Venue} &   \textbf{PSNR}~$\uparrow$~ & \textbf{SSIM}~$\uparrow$& \textbf{Param}~$\downarrow$&  \textbf{RS}  
\\
\shline
 \multicolumn{6}{c}{Image deraining on the UHD-Rain dataset}\\
\shline
UHD & ICCV'21&18.04 & 0.811 & 34.5M& \XSolidBrush\\
Uformer& CVPR'22 & 23.73 & 0.871 & 20.6M& \CheckmarkBold   \\
SFNet&ICLR’23 & 23.64 &0.846 & 13.3M & \CheckmarkBold  \\
\textbf{SimpleIR (Ours)}&-& \textbf{36.88}& \textbf{0.980} & \textbf{4.6M} & \XSolidBrush  \\\shline
\end{tabular}
\caption{Image deraining. 
SimpleIR with at least fewer parameters significantly advances state-of-the-art.
}
\label{tab:Image deraining.} 
\end{table}

\begin{table}[!t]\footnotesize
\tablestyle{2.2pt}{1}
\begin{tabular}{l|c|ccccc}
\shline
 \textbf{Method} &\textbf{Venue} &   \textbf{PSNR}~~ & \textbf{SSIM}~~ & \textbf{Param}~$\downarrow$&  \textbf{RS}  
\\
\shline
 \multicolumn{6}{c}{Low-light image enhancement on UHD-LL dataset.}\\
\shline
Uformer&CVPR'22&21.30  & 0.823  & 20.6M &\CheckmarkBold    \\
LLformer&AAAI'23&24.06 & 0.858 & 13.2M & \CheckmarkBold\\
\textbf{UHDFour}&ICLR'23&\textbf{26.23} & \textbf{0.900} & 17.5M & \XSolidBrush  \\
SimpleIR (Ours)&-&22.66 & 0.891 & \textbf{4.6M} &\XSolidBrush   \\
\shline
\end{tabular}
\caption{Low-light image enhancement. Our SimpleIR achieves impressive results with few parameters.
}
\label{tab:Low-light image enhancement.} 
\vspace{-4mm}
\end{table}

\begin{table}[!t]\footnotesize
\begin{center}
\caption{Comparison of our method with other unified image restoration approaches on low resolution.}
\label{table:standard}
\vspace{-3mm}
\setlength{\tabcolsep}{2pt}
\scalebox{1}{
\begin{tabular}{l c | c | c }
\toprule[0.15em]
 & \textbf{Rain100L} & \textbf{GoPro} & \textbf{BSD ($\sigma$=$50$)} \\
 \textbf{Method~} & \textbf{PSNR}~\textbf{SSIM} &\textbf{ PSNR}~\textbf{SSIM} & \textbf{PSNR}~\textbf{SSIM}\\
\midrule[0.15em]
NAFNet  &~30.46~0.926 & 26.67 0.805 &  28.10 0.805 \\
Restormer &~31.46~0.904 & 25.21~0.752  & 28.59~0.900 \\
PrompIR &~33.97~0.938 & 26.82~0.819 & 29.89~0.824\\
AirNet &~30.99~0.929& 26.50~0.800 & 29.10~0.803 \\
SimpleIR (Ours) &30.89~0.9164 & 27.46~0.8298 & 32.84~0.910 \\
\bottomrule[0.1em]
\end{tabular}
}
\end{center}\vspace{-4mm}
\end{table}
\subsubsection{Results on Multi-task Image Restoration.\\}
We compare with state-of-the-art methods: the CNN-based unified image restoration pipeline NAFNet \cite{chen2022simple} and AirNet \cite{9879292}, the Transformer-based Restormer \cite{9878962} and PromptIR \cite{potlapalli2023promptir}.
They are all fine-tuned for all four tasks. Table.~\ref{tab:Comparison_all} presents quantitative results of SimpleIR and other methods in all 4 tasks.
As demonstrated, SimpleIR achieves competitive performance across all four tasks. 
\setlength{\tabcolsep}{3.5pt}
\begin{table*}[!t]
    \setlength{\abovecaptionskip}{0cm}
    \setlength{\belowcaptionskip}{0cm}
    \caption{Quantitative comparison on desnowing, deraining, deblurring, and low light enhancement tasks. The best results are marked in boldface.}
    \label{tab:Comparison_all}
    \centering
    \footnotesize
    \def\arraystretch{1.3}
    \scalebox{1}{
    \begin{tabular}{cccccccccccccccc}\toprule
        \multirow{3}{*}{\textbf{Method}} & \multicolumn{3}{c}{\textbf{Desnowing}} & \multicolumn{3}{c}{\textbf{Deraining}} & \multicolumn{3}{c}{\textbf{Deblurring}} & \multicolumn{3}{c}{\textbf{LLIE}} & \multicolumn{3}{c}{\textbf{Average}} \\
        &\scriptsize SSIM$\uparrow$ & \scriptsize PSNR$\uparrow$ & \scriptsize LPIPS$\downarrow$ & \scriptsize SSIM$\uparrow$ & \scriptsize PSNR$\uparrow$ & \scriptsize LPIPS$\downarrow$ & \scriptsize SSIM$\uparrow$ & \scriptsize PSNR$\uparrow$ & \scriptsize LPIPS$\downarrow$ & \scriptsize SSIM & \scriptsize PSNR & \scriptsize LPIPS$\downarrow$ & \scriptsize SSIM$\uparrow$ & \scriptsize PSNR$\uparrow$ & \scriptsize LPIPS$\downarrow$  \\
        \cmidrule(l){1-1}  \cmidrule(l){2-4} \cmidrule(l){5-7}  \cmidrule(l){8-10} \cmidrule(l){11-13} \cmidrule(l){14-16}
        NAFNet & 0.927 & 23.41 & 0.188 & 0.801 & 20.86 & 0.396 & 0.818 & 27.36 & 0.239 & \textbf{0.889} & \textbf{22.56} & 0.340 & 0.858 & 23.55 & 0.297  \\
        Restormer & 0.940 & 23.65 & 0.151 & 0.827 & 20.26 & 0.353 & 0.879 & 28.88 & 0.161 & 0.871 & 22.43 & 0.439 & 0.879  & 23.81 & 0.277 \\
        PromptIR & 0.913 & 23.44 & 0.117 & 0.805 & 20.50 & 0.297 & 0.824 & 25.23 & 0.169 & 0.888 & 22.21 & 0.322 & 0.857 & 22.85 & 0.226  \\
        AirNet & 0.906 & 21.95 & 0.178 & 0.786 & 19.46 & 0.383 & 0.809 & 22.06 & 0.247 & 0.805 & 16.97 & 0.416 & 0.826  & 20.11 & 0.306\\
        SimpleIR (Ours) & \textbf{0.983} & \textbf{28.29} & \textbf{0.013} & \textbf{0.976} & \textbf{36.31} & \textbf{0.024} & \textbf{0.868} & \textbf{30.37} & \textbf{0.150} & 0.886 & 22.16 & \textbf{0.125} & \textbf{0.928} & \textbf{29.28} & \textbf{0.077}\\
        \bottomrule
    \end{tabular}
    }
    \vspace{-4mm}
\end{table*}
\vspace{-2mm}
\section{Ablation Study}
In this section, we focus on the effects of the review quantity on the model performance.
\begin{table*}[!t]\footnotesize
\setlength{\abovecaptionskip}{0cm}
    \setlength{\belowcaptionskip}{0cm}
    \caption{Quantitative comparison on desnowing, deraining, deblurring, and low light enhancement tasks. The best results are marked in boldface and the second-best results are underlined.}
    \label{tab:ablation_study_1} 
    \centering
    \footnotesize
    \def\arraystretch{1.3}
\scalebox{1}{
\begin{tabular}{ccccccccccc}
\shline
 \multirow{2}{*}{\textbf{Review Quantity}} &\multicolumn{2}{c}{\textbf{Desnowing}}~~  &   \multicolumn{2}{c}{\textbf{Deraining}}~~ & \multicolumn{2}{c}{\textbf{Deblurring}}~~ & \multicolumn{2}{c}{\textbf{LLIE}}~~ &  \multicolumn{2}{c}{\textbf{Average}}~~   
\\
& \scriptsize SSIM& \scriptsize PSNR & \scriptsize SSIM & \scriptsize PSNR & \scriptsize SSIM & \scriptsize PSNR & \scriptsize SSIM & \scriptsize PSNR & \scriptsize SSIM & \scriptsize PSNR \\
\shline
0\%& 0.902 & 25.13 & 0.899 & 30.92 & 0.800 & 25.77 & 0.890 & 22.55 &  0.873 & 26.09 \\
10\%& 0.920 & 26.29 & 0.901 & 32.31 & 0.838 & 27.37 & 0.886 & 21.26 & 0.886 & 26.81  \\
40\%& 0.971 & 27.33 & 0.966 & 35.89 & 0.843 & 28.77 & 0.878 & 20.11 & 0.914 & 27.78 \\
20\% (Ours)& \textbf{0.983} & \textbf{28.29} & \textbf{0.976} & \textbf{36.31} & \textbf{0.868} & \textbf{30.37} & \textbf{0.886} & \textbf{22.16} & \textbf{0.928} & \textbf{29.28}  \\
\shline
\end{tabular}
}
\vspace{-4mm}
\end{table*}
\par To ascertain the optimal review quantity, we conducted an ablation study varying the review quantity.
We modify the review ratio while keeping all other training parameters constant. The model underwent training with the following review ratios: 0\%, 10\%, 40\%. For each setting, we record the performance metrics after the completion of the LLIE training phase.
\par
\begin{figure}[htbp]\footnotesize
\centering
\scalebox{0.85}{
    \includegraphics[width=\linewidth]{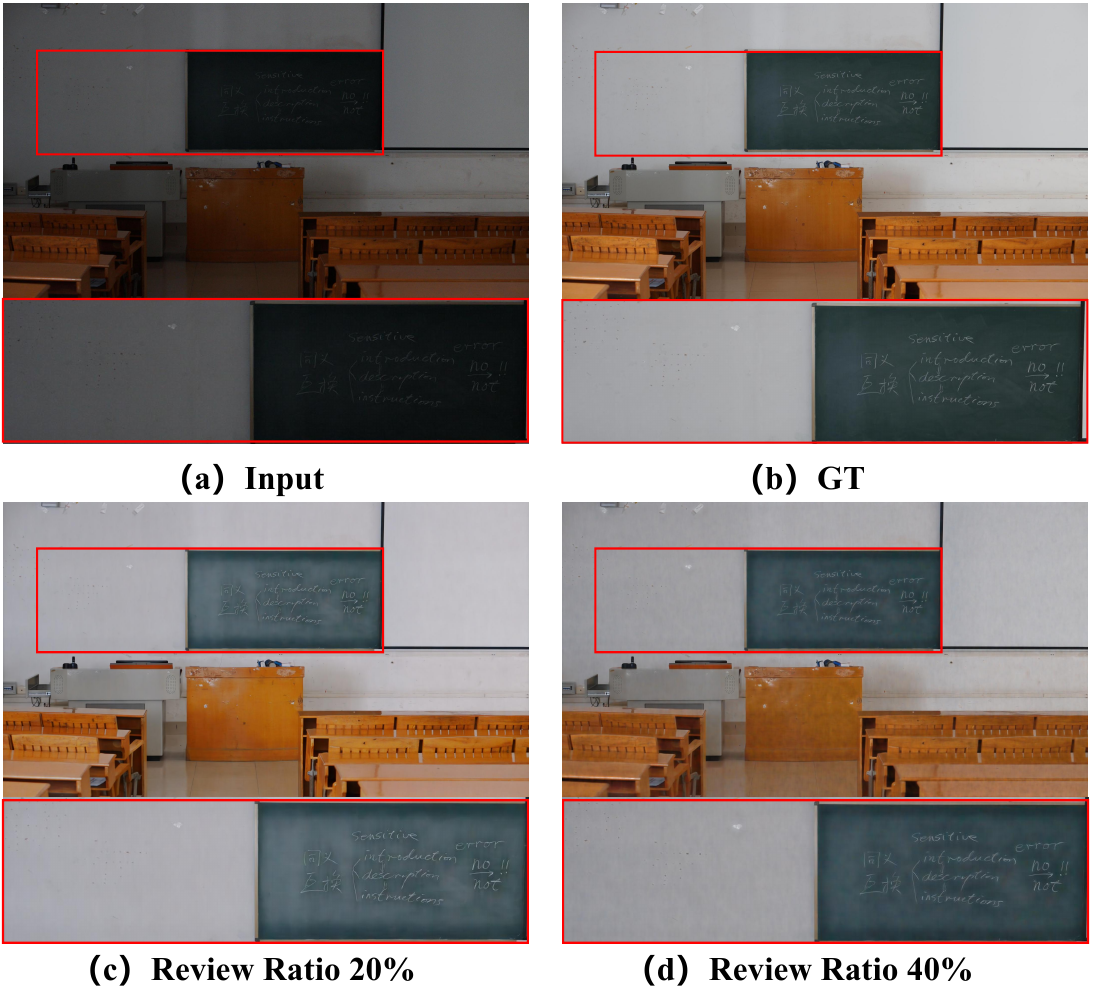}}
\vspace{-2mm}
\caption{Comparision of different review ratios. A high review ratio leads to a speckled background.}
\label{fig:fail}
\vspace{-4mm}
\end{figure}

 As demonstrated in Table~\ref{tab:ablation_study_1}, model performance is sensitive to review data quantity. Without review, the model's baseline performance indicates limited generalization ability. A 10\% review ratio yields a modest improvement, implying that moderate review can reinforce knowledge effectively.
However, when the review ratio is further increased to 40\%, the model's performance does not show a proportional increase. 
As shown in Figure~\ref{fig:fail}(d), we notice that the background of the restored image exhibits a mottled background. The overabundance of reviewed difficult cases could lead to information confusion, thereby impacting the model's ability to restore the image context accurately.
\vspace{-2.6mm}
\section{Discussion and Analysis}
We evaluate them on four degradation-specific tasks: image deraining on the Rain100L dataset (Yang et al.~\shortcite{Jiang_2020_CVPR}), image deblurring on the GoPro dataset (Nah et al.~\shortcite{gopro2017}), and image denoising on BSD (Martin et al.\shortcite{martin2001database}). 
Table.~\ref{table:standard} shows that our SimpleIR achieves consistent performance.
We provide several visual examples for each task for a better understanding of the various degradations and datasets, as shown in Figure \ref{fig:low-resolution}.
We also train the SimpleIR by random learning trajectory, the result report in the appendix.
\begin{figure}[htbp]\footnotesize
\centering
\scalebox{0.84}{
    \includegraphics[width=\linewidth]{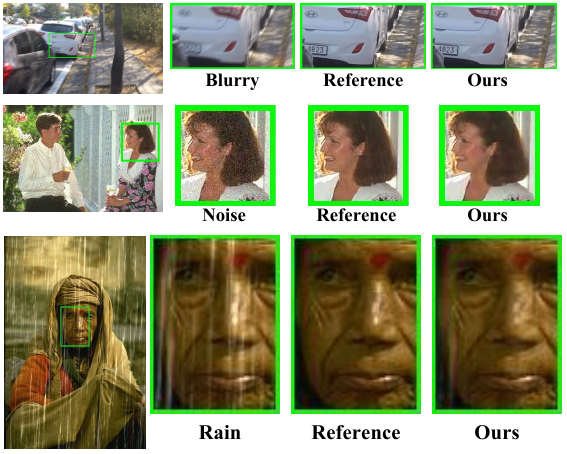}}
\vspace{-3.5mm}
\caption{Example images with 3 image restoration tasks. For each task, the first column is the corrupted input and the last column is the result produced by our unified image restoration model.}
\label{fig:low-resolution}
\vspace{-4mm}
\end{figure}
\vspace{-4mm}
\section{Conclusion}
In this paper, we propose a training paradigm, Review Learning to unify UHD image restoration tasks.
In contrast to previous research that required customized network architecture and design prompts, review learning can help arbitrary image restoration models achieve the ability to cope with a wide range of degraded images.
It only requires sequential training according to the intrinsic task complexity and periodic review of the challenging samples.
In addition, our framework can efficiently reason about degraded images with 4K resolution on a single consumer-grade GPU.
\bibliography{cite}
\newpage
\twocolumn[{%
\renewcommand\twocolumn[1][]{#1}%
\section{\Huge\textbf{Supplementary Material}}
\begin{center}\footnotesize
    \centering
    \captionsetup{type=figure}
	\includegraphics[width=0.85\linewidth]{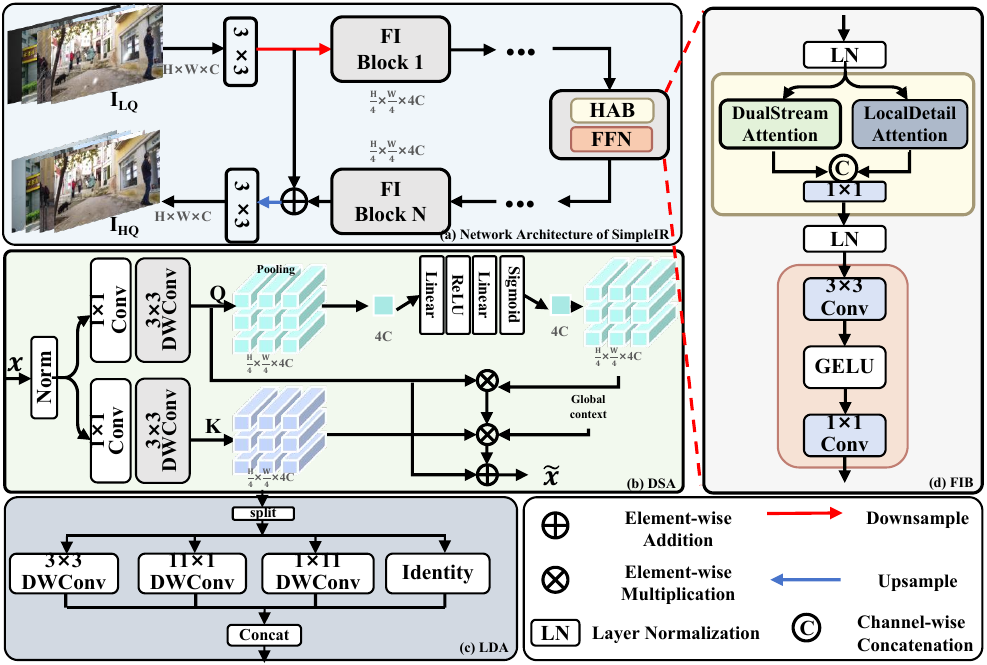}
    \caption{Overview of the \textbf{SimpleIR} framework. SimplIR first transforms the input LR image into the feature space using a convolutional layer and a downsampler module, performs feature extraction using a series of feature iteration blocks (FIBs), and then reconstructs these extracted features by an upsampler module. The FIB is mainly implemented by a hybrid attention block (HAB) and a feed-forward network (FFN).}
    \label{fig:framework}
\end{center}%
}]
In this supplementary material, we first provide more details of the framework in Sec. \textcolor{blue}{\textbf{A}}. Then, we show several visual examples for each task for a better understanding of the degradations and datasets in Sec. \textcolor{blue}{\textbf{B}}. 
%
%
In addition, we select a random learning trajectory and report the training outcomes in comparison to those obtained by following a curriculum that gradually increases task difficulty and reports the result in Sec. \textcolor{blue}{\textbf{C}}.
Finally, to better illustrate the effectiveness of our method in the real scene, we qualitatively compare our proposed SimpleIR with the Restormer and NAFNet methods in real-world images and implement object detection experiments in Sec. \textcolor{blue}{\textbf{D}}.
\section{\textbf{A~~}Overview of the SimpleIR Approach}
\label{sec: approach}
In this section, we give more details about SimpleIR, shown in Figure~\ref{fig:framework}.
Especially, since loss serves as a criterion for selecting challenging samples in our review learning process, we adopt the approach that utilizes the loss $L_{1}$ and an FFT-based frequency loss function, which is defined as:
\begin{equation}
    \mathcal{L}=\|I_{LQ}-I_{HQ}\|_1+\lambda\|\mathcal{F}(I_{LQ})-\mathcal{F}(I_{HQ})\|_1
\end{equation}
where $\|\cdot\|_{1}$ denotes $L_{1}$ norm; $\mathcal{F}$ denotes the Fast Fourier transform; 
$\lambda$ is the weight that is empirically set to be 0.1.
\begin{figure*}[!t]
\setlength{\belowcaptionskip}{-0.2cm}
	\centering
	\includegraphics[width=\linewidth]{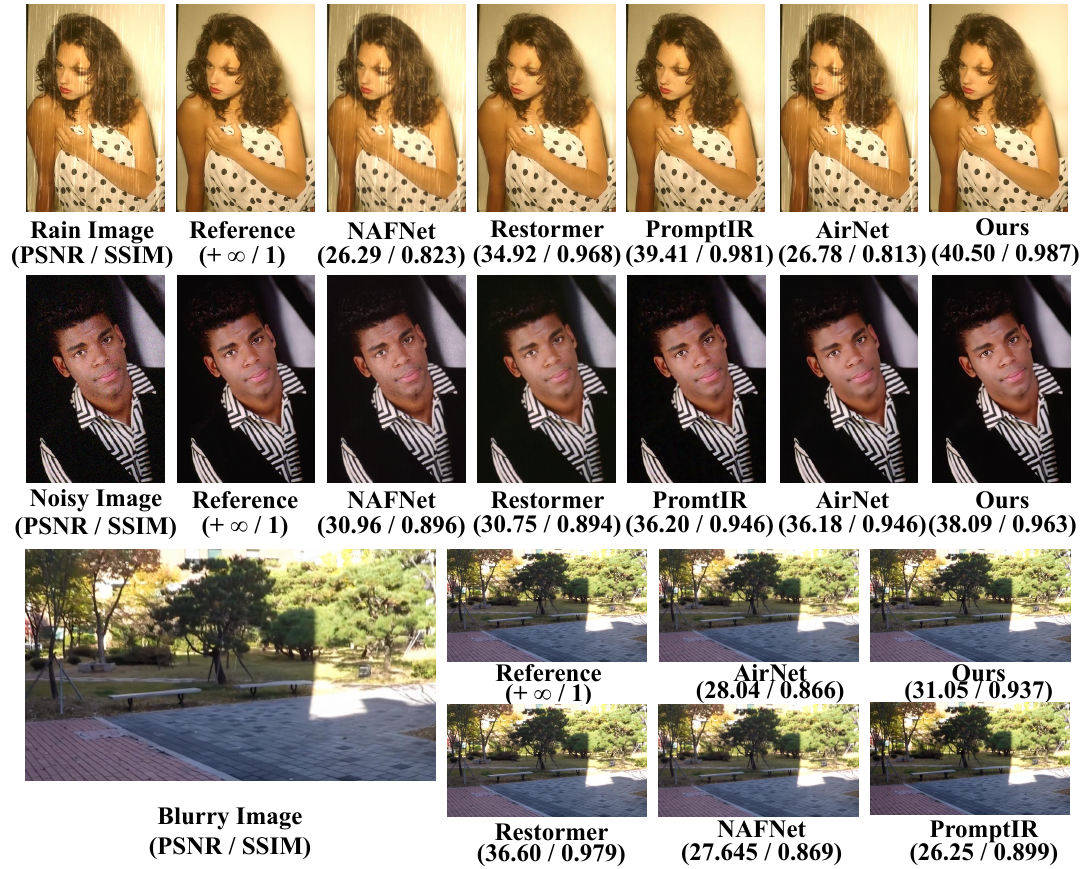}\\
	\caption{Comparison of our method with other approaches on the unified image restoration.}
	\label{fig:low-resolution}
\end{figure*}

\section{\textbf{B~~}More Results about Image Restoration}\label{sec: Visual}
\begin{figure*}[!t]
\setlength{\belowcaptionskip}{-3.5cm}
	\centering
	\includegraphics[width=\linewidth]{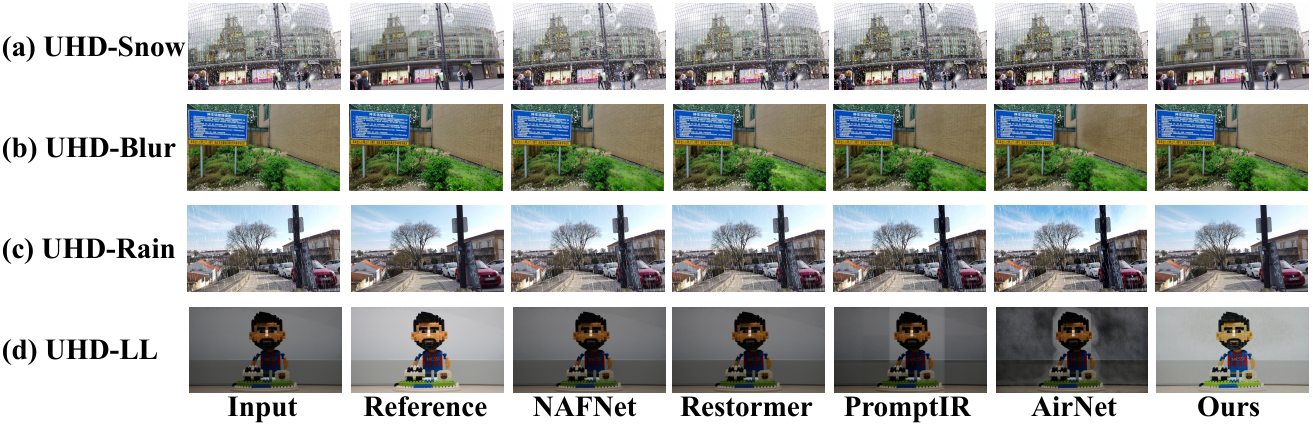}\\
	\caption{Comparison of our method with other approaches on the unified image restoration on UHD images.}
	\label{fig:uhd}
\end{figure*}

In this section, we present various degradation datasets restoration results, as shown in Figure~\ref{fig:low-resolution} and Figure~\ref{fig:uhd}. 
\section{\textbf{C~~}Random Learning Trajectory Results}
\label{sec: Random}
In this section, we select the training sequence randomly: UHD-Rain, UHD-Snow, UHD-LL, UHD-Blur. In addition, we present the result in Table \ref{tab: random}.
The random learning trajectory appears to result in slightly lower performance metrics across the board. This suggests that the lack of a gradual difficulty curve may hinder the model's ability to fully capitalize on the learning opportunities presented by each task.
The UHD-LLIE task, in particular, exhibits a significant performance gap, with a PSNR of 15.10, SSIM of 0.771, and LPIPS of 0.445 for the random trajectory, compared to 22.16, 0.886, and 0.124 respectively for the structured path. This disparity underscores the importance of a curriculum that aligns with the complexity of the tasks, potentially allowing the model to build a stronger foundation before tackling more challenging scenarios.
This underscores the hypothesis that a structured approach, which incrementally increases in difficulty, can be more conducive to achieving higher performance in task-specific learning scenarios.
\begin{table}[hbpt]\footnotesize
\centering
\begin{tabular}{lccc}
\hline
\textbf{Task} & \textbf{PSNR} & \textbf{SSIM} & \textbf{LPIPS} \\
\hline
\multicolumn{4}{c}{\textbf{Random Learning Trajectory Results}} \\
\hline
UHD-Snow & 27.40 & 0.981 & 0.039 \\
UHD-Rain & 34.55 & 0.963 & 0.120 \\
UHD-Blur & 24.66 & 0.800 & 0.230 \\
UHD-LLIE & 15.10 & 0.771 & 0.445 \\
\hline
\end{tabular}
\captionof{table}{Random Learning Trajectory Results}
\label{tab: random}
\end{table}
\begin{table}[hbpt]\footnotesize
\centering
\begin{tabular}{lccc}
\hline
\centering
\textbf{Task} & \textbf{PSNR} & \textbf{SSIM} & \textbf{LPIPS} \\
\hline
\multicolumn{4}{c}{\textbf{Training by Learning Path Results}} \\
\hline
UHD-Snow & 28.29 & 0.983 & 0.013 \\
UHD-Rain & 36.31 & 0.976 & 0.024 \\
UHD-Blur & 30.37 & 0.868 & 0.150 \\
UHD-LLIE & 22.16 & 0.886 & 0.124 \\
\hline
\end{tabular}
\captionof{table}{Training by Learning Path Results}
\label{tab: learningpath}
\end{table}
\section{\textbf{D~~}Results on Real-world Images} \label{sec: real-world}
\begin{figure*}[!t]
	\begin{center}
	    \includegraphics[width=\linewidth]{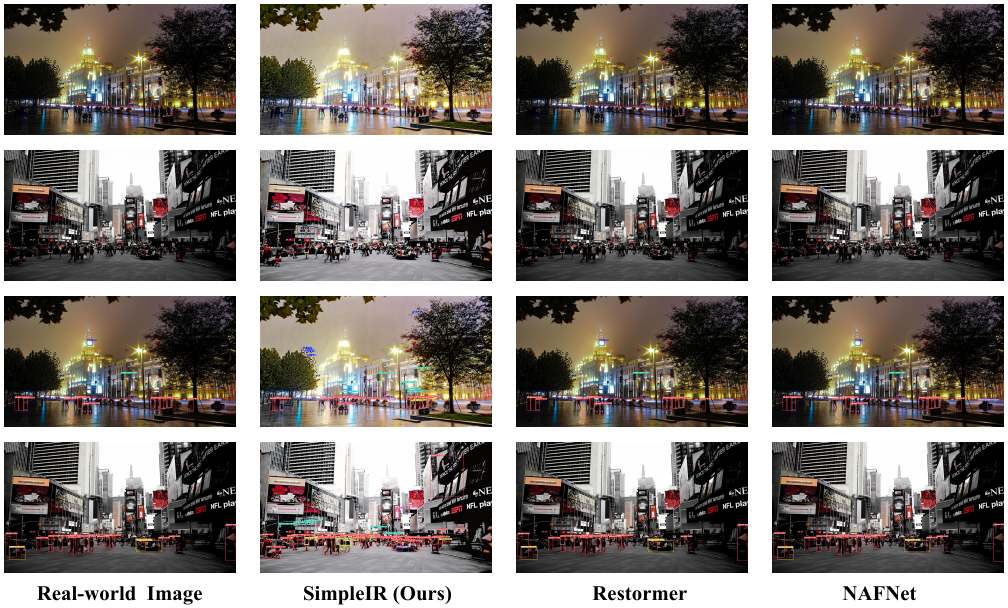}
	\end{center}
	\caption{Visual comparison on the real-world images. The proposed method produces visually more pleasing results. The last two rows are the object detect results by YOLOV$5$. } 
	\label{fig:real-world}
\end{figure*}

Figure \ref{fig:real-world} displays the visual comparison results. 
The comparison results of a real-world image further demonstrate the superiority of our proposed method.
The object detection result can be observed when low-light images are processed with our SimpleIR before performing object detection, more objects can be detected.

\end{document}